\RequirePackage{fix-cm}
\documentclass[twocolumn]{svjour3}          % twocolumn

\usepackage{multirow}
\usepackage{graphicx, epstopdf}
\usepackage{xspace,epsfig,url}
\usepackage{amssymb,amsmath}
\usepackage[noadjust]{cite}
\usepackage{enumerate}
\usepackage{float}
\usepackage{amsmath}
\usepackage{epsf}
\usepackage{psfrag}
\usepackage{verbatim}
\usepackage[all]{xy}
\usepackage{color}
\usepackage{comment}
\usepackage{subfigure}
\usepackage{cases}

\newcommand{\vect}[1]{\mathbf{#1}}
\newcommand{\vects}[1]{\boldsymbol{#1}}

\graphicspath{{./Figures/}}

\begin{document}

\title{\LARGE \bf Design and Kinematic Optimization of a Novel Underactuated Robotic Hand Exoskeleton}

\author{Mine Sarac \and
        Massimiliano Solazzi \and
        Edoardo Sotgiu \and Massimo Bergamasco \and
        Antonio Frisoli
}

\authorrunning{Sarac et al} % if too long for running head
\institute{Mine Sarac, Massimiliano Solazzi, Edoardo Sotgiu, Massimo Bergamasco, Antonio Frisoli\at
              PERCRO Lab, Scuola Superiore Sant'Anna, Pisa, Italy. \\
              \email{m.sarac, m.solazzi, e.sotgiu, m.bergamasco, a.frisoli@sssup.it}
}

\maketitle
\thispagestyle{empty}
\pagestyle{empty}

\begin{abstract}
This study presents the design and the kinematic optimization of a novel, underactuated, linkage-based robotic hand exoskeleton to assist users performing grasping tasks. The device has been designed to apply only normal forces to the finger phalanges during flexion/extension of the fingers, while providing automatic adaptability for different finger sizes. Thus, the easiness of the attachment to the user's fingers and better comfort have been ensured. The analyses of the device kinematic pose, statics and stability of grasp  have been performed. These analyses have been used to optimize the link lengths of the mechanism, ensuring that a reasonable range of motion is satisfied while maximizing the force transmission on the finger joints. Finally, the usability of a prototype with   multiple fingers has been tested during grasping tasks with different objects.
\end{abstract}


\section{Introduction}

A hand exoskeleton is a wearable haptic device  providing  haptic feedback in virtual environments or  motor assistance for robotic assisted rehabilitation to the user's hand.

The design criteria adopted for hand exoskeletons change according to the application. For instance, glove-based devices~\cite{Kobayashi2012, Lee2013} can guide the human fingers in a natural manner and control all finger joints efficiently, however the patients with disabilities might face difficulties in wearing the device. On the contrary,  devices~\cite{Brokaw2011, Aubin2013} coupling all fingers are designed to perform repetitive movements for the rehabilitation therapies by attaching all fingers together to perform the same movement, however they do not allow individual finger movement and  make the device impractical during assistance to real grasping tasks.

The linkage-based devices might be efficiently used for  physical rehabilitation thanks to their independence of the fingers and the ease of wearability. Such devices can be categorized further based on the number of actuators for each finger component.
From the kinematic point of view, human fingers with the exception of the thumb have $4$ Degrees-of-Freedom (DoFs): $3$ flexion/extension and $1$ abduction/adduction joints.
 The devices that control all the finger joints with independent actuators provide full mobility during the grasping tasks~\cite{Li2011, Hasegawa2011, Jones2012}. Although linkage-based devices can control all the finger joints individually allows the full posture control of the finger, however, they mostly suffer from the heavy and high-cost design, while sacrificing the portability.

%Even though designing a haptic device to control all $4$ DoFs is feasible~\cite{Wang2009, Li2011}, focusing on the flexion/extension joints might improve the simplicity and the portability during operation.

In alternative one single actuator can be used to control the fingertip position from a single contact point \cite{Iqbal2011}. However these devices cannot provide individual joint rehabilitation and implicit posture control of the finger joints. Moreover, these devices move the finger in a predefined path to reach the desired position by the fingertip, so they might suffer from the lack of adaptability of the shape and the size of the grasping objects as well as the accuracy and safety of the performed task for the finger. Nevertheless, they have portable, easily wearable, light and low-cost design. The mechanical system can also be designed with a differential system to transmit forces to multiple finger phalanges from a single actuator \cite{Tang2013, Taheri2014, Chiri2012}. The mechanical adjustments on the system might change the relation between the finger joints. Even though this approach makes the device lighter and lower cost than the ones with the multiple actuators, the complexity of the design remains.

Another way to implement a single actuator hand exoskeleton is to embrace underactuation~\cite{Gosselin2003} by introducing extra DoFs to the mechanism through passive joints or elastic elements. The underactuation based finger exoskeleton~\cite{Ertas2014} can imitate the natural finger movements during  therapy while improving the wearability, the portability and the adjustability for the grasping object automatically. The grasping tasks are performed through the position control of the actuator while the forces are transmitted to the finger phalanges based on the contact forces. Although  the complete posture control of the finger joints cannot be achieved through underactuation, however if the human finger is considered as  part of the mechanism,  its intrinsic finger joints mechanical impedance  can replace the role of  additional springs or elastic elements used in underactuated grippers \cite{birglen2004kinetostatic}.

\noindent The use of rotational actuators in this device might prevent the extension of the finger exoskeleton to multiple fingers while maintaining the portability of the current prototype. Even though the force transmission is suggested to be natural during operation, there is no implication regarding the direction of applied forces to the finger phalanges.

Table~\ref{tab:intro} provides a summary regarding some of the devices in the literature, comparing them in terms of the use of single actuator, the automatic adaptability for different finger sizes of the users, the automatic object adaptability with respect to the size or shape and the implementation of posture control strategies through independent joint control.

\begin{table}[htb]
   % \vspace*{-1\baselineskip}
%\footnotesize
\caption{Comparison between existing studies based on using a single actuator, adapting the operation for finger size, adapting the operation for the object to encounter, and posture control for user's hand.}
\label{tab:intro}
\begin{center}
\resizebox{3.1in}{!}{
\begin{tabular}{p{1in}|| p{0.5in}| p{0.6in}| p{0.5in}| p{0.45in}}
  %\hline
  \hline \hline \\
  Ref.  & Single Actuator  & Finger Size Adapt. & Object Adapt. & Posture Control\\
  \hline
 Li et al.~\cite{Li2011} & x & x &  & x \\ [1ex] %\hline
 Hasegawa et al.~\cite{Hasegawa2011} &   & & & x\\ [1ex] %\hline
 Jones et al.~\cite{Jones2012} & & & & x\\ [1ex] %\hline
 Chiri et al.~\cite{Chiri2012} & x & & & x\\ [1ex] %\hline
 Iqbal et al.~\cite{Iqbal2011} & x & & & \\ [1ex] %\hline
 Tang et al.~\cite{Tang2013} & x & x & & \\ [1ex] %\hline
 Taheri et al.~\cite{Taheri2014} & x & & &  \\ [1ex] %\hline
 Ertas et al.~\cite{Ertas2014} & x & x & x &  \\ [1ex]
 \hline \hline
\end{tabular}}
\vspace*{-2\baselineskip}
\end{center}
\end{table}

In this paper, we propose a novel underactuated hand exoskeleton to ensure that the transmitted forces between the device and the user are always perpendicular to the finger phalanges in order to improve the efficiency and naturality of the finger movements during operation. Each linkage-based finger component is placed on top of the finger and run by a single linear actuator. The underactuation approach has been adopted in order to achieve automatic adjustment of the grasping task with various shapes and sizes as much as the hand size of the user. The linear actuators can be fitted on top of the hand by improving the portability of the overall system. By considering the human finger as integral  part of the mechanism, the role of additional springs or elastic elements  required in underactuation mechanisms is performed by the  intrinsic mechanical impedance capabilities of human finger joints.
 Thanks to the improved wearability and the capability for real grasping tasks, the device can be used for rehabilitation tasks, as well as haptic applications. The link lengths of each finger component have been optimized to achieve the highest performance in terms of the force transmission while covering a wide workspace of finger joints. The optimized multi-finger hand exoskeleton has been tested for feasibility for the real grasping tasks.

The paper continues as follows: Section \ref{sec:design} presents the discussion on the design requirements and the kinematic synthesis of the proposed mechanism. Furthermore, Subsection \ref{sec:kinematics} and Subsection \ref{sec:statics} detail the kinematics and statics analyses respectively. Optimization procedure to decide the exoskeleton link lengths is presented in Section \ref{sec:optimization}, defining the sensitivity analysis, kinematic and static constraints and the optimization objective. Section \ref{sec:hand} defines the implementation of the finger exoskeleton for the index finger and for multi fingers. Finally, Section \ref{sec:conc} concludes the paper, discussing the future works.

\section{Design Requirements \& Device Characteristics} \label{sec:design}

\begin{figure}[h]
  \centering
      %\vspace*{-.5\baselineskip}
  \resizebox{3in}{!}{\includegraphics{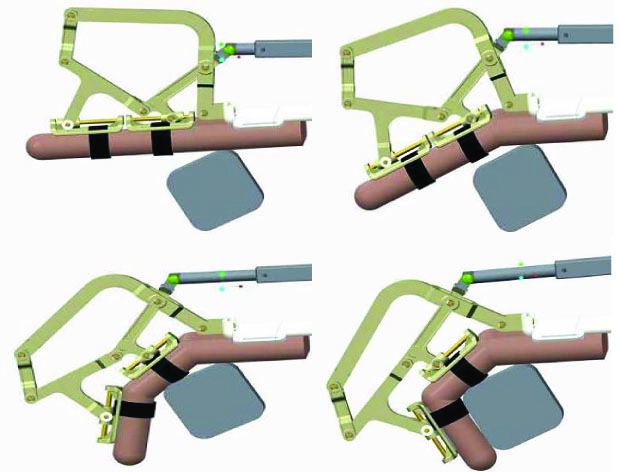}}
        \caption{How the underactuation concept works: the exoskeleton moves the first finger joint, until the first finger phalange reaches a contact. Then, the actuation is transmitted to the next joint automatically.}
        \label{fig:underact}
   % \vspace*{-1\baselineskip}
\end{figure}

Except the thumb, human fingers can be modelled as a planar kinematic open chain with $3$ joints: distal interphalangeal (DIP), proximal interphalangeal (PIP), and metacarpophalangeal (MCP) from the fingertip to the palm of the hand. The MCP joint acts like a spherical joint with flexion/extension and abduction/adduction movements, while PIP and DIP joints perform only flexion/extension. The mechanical design to assist the grasping tasks can be simplified and restricted to a planar scheme by neglecting the abduction/adduction movement as can be seen in Fig.~\ref{fig:underact}. The natural range of motion (RoM) for the corresponding joints~\cite{Wang2009} during flexion/extension are presented in Table~\ref{tab:index_rom}.

\begin{table}[htb]
     \vspace*{-1\baselineskip}
\caption{Ranges of motion for finger joints.}
\label{tab:index_rom}
\begin{center}
\begin{tabular}{c||c|c|c}
\hline \hline \\
  Joint & MCP & PIP & DIP \\\hline
  RoM & 0 - 85 & 0 - 100 & 0 - 80 \\
  \hline \hline
\end{tabular}
\end{center}
    \vspace*{-2\baselineskip}
\end{table}

The proposed hand exoskeleton provides assistance only to the MCP and PIP joints to improve the simplicity and the wearability during real grasping tasks. Such restriction can be ensured only by connecting the device to the first and the second phalanges of the human finger and leaving the third phalange free. The safety of the operation requires the controlled joints to be consistent with the natural RoM stated previously in Table~\ref{tab:index_rom}. A generic device for various application alternatives might require the following properties to be satisfied:

\begin{itemize}
    \item Comfortable and easy wearability,
   	\item Self-adjustability to different sizes of the fingers,
	\item Effective transmission of forces to the phalanges, and
	\item Grasping of objects with generic shape.
\end{itemize}

To match the given design requirements, the following assumptions have been made for guiding the design of the device:

\begin{description}
    \item[ {\bf Self-adaptability to hand size.}] In order to achieve an adaptable kinematic design for different hand sizes without any mechanical adjustment, the finger phalanges are considered as a part of the kinematic chain~\cite{leonardis2015emg}.
    \item[ {\bf Allowed misalignment of joints.}] The device joints do not need to be aligned with the finger joints.
    \item[ {\bf Effective transmission forces to phalanges.}] Only normal forces are applied to the user's finger during operation. This property significantly improves the design and the functionality of the fasteners, which attach the exoskeleton to the fingers. In particular, since the fasteners are crucially used to transmit the tangential shear forces, there is no need to excessively tighten the finger for transmitting either torques or longitudinal forces to the skin.
    \item[ {\bf Underactuation.}] In order to simplify the number of actuators, underactuation~\cite{Gosselin2003} has been adopted for each finger component by transmitting forces based on the contact forces on the phalanges. This property allows the device to apply stable forces to the finger phalanges during grasping tasks using objects with any shape and size. Moreover, the use of single actuator maintains the lightweight and compact design of the device.
    \item[ {\bf Dorsal structure}] The device is placed above the fingers in order to keep the palm and the sides of the hand free for real grasping tasks and for the multi-finger implementation without mechanical interference.
\end{description}

\begin{figure}[h]
  \centering
      \vspace*{-2\baselineskip}
  \resizebox{3in}{!}{\includegraphics{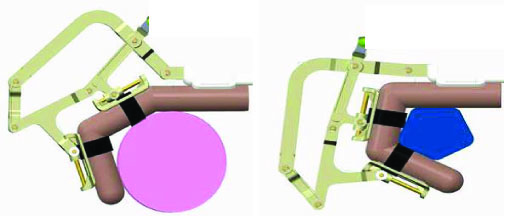}}
        \caption{An underactuated hand exoskeleton can assist user's finger to grasp objects with different sizes without any mechanical adjustments. }
        \label{fig:underact2}
    %\vspace*{-1\baselineskip}
\end{figure}

The force transmission of an underactuated mechanism is based on the contact forces during the grasping tasks to adjust the device for the objects with different sizes and shapes automatically as in Fig.~\ref{fig:underact2}. Even though the underactuation concept does not have the ability to fully constrain a given posture control of the fingers, the adopted design guarantees that the exchanged forces between the fingers and the device are always direct to opening/closing the hand.

In particular, Fig.~\ref{fig:underact} presents the operation flow of a grasping task with the assistance of the underactuated exoskeleton for a single finger. In the beginning of the task, the actuated device moves the MCP joint of the human finger until the proximal phalange reaches and touches the grasping object. When the motion around the MCP joint is constrained due to the contact forces on the first phalange, the flexion is transferred to the PIP joint until the intermediate phalange reaches the grasping object to satisfy the second connection and the grasping task. Similarly, the extension of the finger starts from the PIP joint and is transmitted to the MCP joint until the finger is totally extended and the finger joints reach their physical limits.

The device has been designed to apply only normal forces to the finger phalanges during operation. To achieve this property, the connection links have been mounted to the phalanges through a cylindrical joint and a rotational joint with perpendicular axis (see Fig.~\ref{fig:definitions}), so that the longitudinal forces and the torques are prevented to be applied on the finger phalanges by the mechanism. The absence of the longitudinal forces allows the finger to be connected to the exoskeleton by simple straps around the phalanges.

\section{Pose Analysis} \label{sec:kinematics}

\begin{figure}[htb]
	\centering
	\vspace*{-1.5\baselineskip}
	\subfigure[Kinematic scheme]{\includegraphics[width=0.4\textwidth]{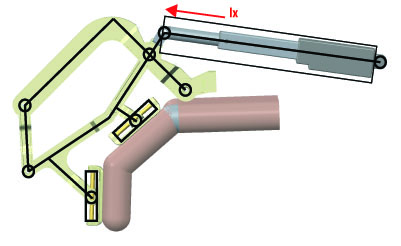}}\\
	\subfigure[CAD model with notation]{\includegraphics[width=0.4\textwidth]{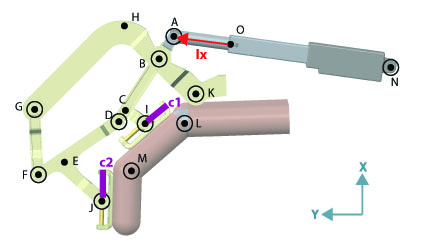}} \\
	\subfigure[CAD model with joint definitions]{\includegraphics[width=0.4\textwidth]{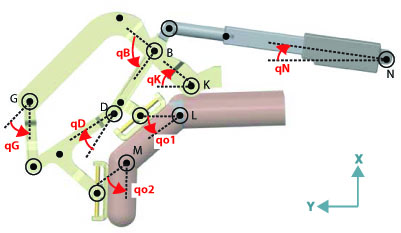}}
	%\vspace*{-0.5\baselineskip}
	\caption{Design concept of a finger component of the underactuated hand exoskeleton: (a) kinematic scheme and joint placements, (b) all necessary points for kinematic analysis, and (c) joint definitions.}
	\label{fig:definitions}
	%\vspace*{-1\baselineskip}
\end{figure}

Fig.~\ref{fig:definitions} shows the design of the proposed linkage-based underactuated mechanism, with the active and passive joints superimposed on the CAD model of the device. In particular, a linear actuator is attached to the point $A$ with the corresponding displacement $l_x$. The MCP and PIP joints of the user's finger are defined as points $L$ and $M$ respectively, with the rotations specified as $q_{o1}$ and $q_{o2}$. The mechanism consists of $9$ passive revolute joints at points $A$, $B$, $D$, $F$, $G$, $I$, $J$, $K$, and $N$ while their rotations are represented as $q_i$ for point $i$. The mechanism is connected to the passive sliders attached to the first and the second phalanges of the user's finger at the points $I$ and $J$, while the displacement of the corresponding passive linear joints are defined as $c_1$ and $c_2$. The finger phalanges are connected to the device at points. The points $H$, $C$ and $E$ are the required points for the pose analysis. The point $N$ presents the point where the actuator is connected to the base. Finally, the point $O$ shows the initial position of point $A$ when the actuation stroke $lx$ is zero.

A pose analysis of the mechanism is required in order to understand the behavior of the mechanical joints, to study the force transmission on the finger phalanges, to avoid the physical limitations of the mechanism, to avoid any mechanical interference during operation and to optimize the system performance. However, the additional DoF, which is introduced by the underactuation concept, prevents the existence of a unique solution for the pose of the finger phalanges only with the given actuator displacement. In other words, it is not possible to analyze the pose of the finger using only the displacement of the actuator. With this motivation, a pose analysis has been performed to obtain a unique configuration for the actuator displacement and the mechanism configuration with the given pose of the finger joints.

The pose analysis has been performed by defining the mechanical closed loops of the system with a set of vector equations. Table~\ref{tab:variables} describes the unknown parameters, constants and finger pose parameters to define the vector between two points stated in Figure~\ref{fig:definitions}. In this table, the notation $\vect{r}^{i}_{j}$ is used to indicate the vectors connecting point $i$ to point $j$, while $q_k$ indicates the angle definition around the point $k$ and $l_{ij}$ shows the constant length between the point $i$ and point $j$. The notations $c_1$, $c_2$, $q_{o1}$ and $q_{o2}$ have been defined to indicate the linear displacements along the finger phalanges and the finger joint rotations. Considering that the pose analysis will be performed for the given finger pose, the finger pose parameters have been specified separately.

\begin{table}[h]
%\vspace*{-1.5\baselineskip}
\caption{Variable and constants for the vectors expressed in Fig.\ref{fig:definitions}.}
\label{tab:variables}
\begin{center}
\resizebox{2.9in}{!}{
\begin{tabular}{p{0.35in}|| p{0.6in}| p{0.6in}| p{0.7in}}
  \hline\hline \\
  Vector  &  Unknown Parameter & Constants & Finger Pose Parameter \\ [1ex] \hline
  $\vect{r}^{O}_{A}$ & \multirow{2}{*}{$l_x, q_N$}   &  & \multirow{6}{*}{}  \\ [1ex] \cline{3-3}
   $\vect{r}^{O}_{N}$ &                              & $l_{act}$ & \\ [1ex] \cline{2-3}
  $\vect{r}^{A}_{D}$ &\multirow{2}{*}{$q_B$}   & $l_{AD}$ & \\ [1ex] \cline{3-3}
   $\vect{r}^{C}_{I}$ &  & $l_{CI}$ &\\ [1ex] \cline{2-3}
  $\vect{r}^{G}_{K}$ & $q_K$ & $l_{GK}$ &  \\ [1ex] \cline{2-3}
   $\vect{r}^{K}_{N}$ &  & $l_{KN}$, $q_{KN}$ &   \\ [1ex]  \cline{2-4}
   $\vect{r}^{I}_{L}$ & \multirow{2}{*}{$c_1$}  &  &\multirow{2}{*}{$q_{o1}$}\\ [1ex] \cline{3-3}
    $\vect{r}^{M}_{I}$ &  & $l_{ML}$ &  \\ [1ex] \cline{2-4}
  $\vect{r}^{L}_{K}$ &  & $l_{LK}$, $q_{LK}$ &\\[1ex] \cline{2-3}
  $\vect{r}^{D}_{F}$ & \multirow{2}{*}{$q_D$}  & $l_{DF}$ & \\ [1ex] \cline{3-4}
  $\vect{r}^{E}_{J}$ &  & $l_{EJ}$ & \\ [1ex] \cline{2-4}
  $\vect{r}^{J}_{M}$ & $c_2$ &  & $q_{o2}$\\ [1ex] \cline{2-4}
  $\vect{r}^{G}_{F}$ & $q_G$ & $l_{GF}$ & \\ [1ex]
\hline \hline
\end{tabular}}
\vspace*{-2\baselineskip}
\end{center}
\label{table_parameters}
\end{table}
%%  $\vect{r}^{O}_{A}$ & $l_x, q_N$ &  & \\ [1ex] \hline
%%$\vect{r}^{O}_{N}$ & $q_N$ & $l_{act}$ & \\ [1ex] \hline

\begin{figure*}[htb]
  \centering
      \vspace*{-2\baselineskip}
  \resizebox{6.5in}{!}{\includegraphics{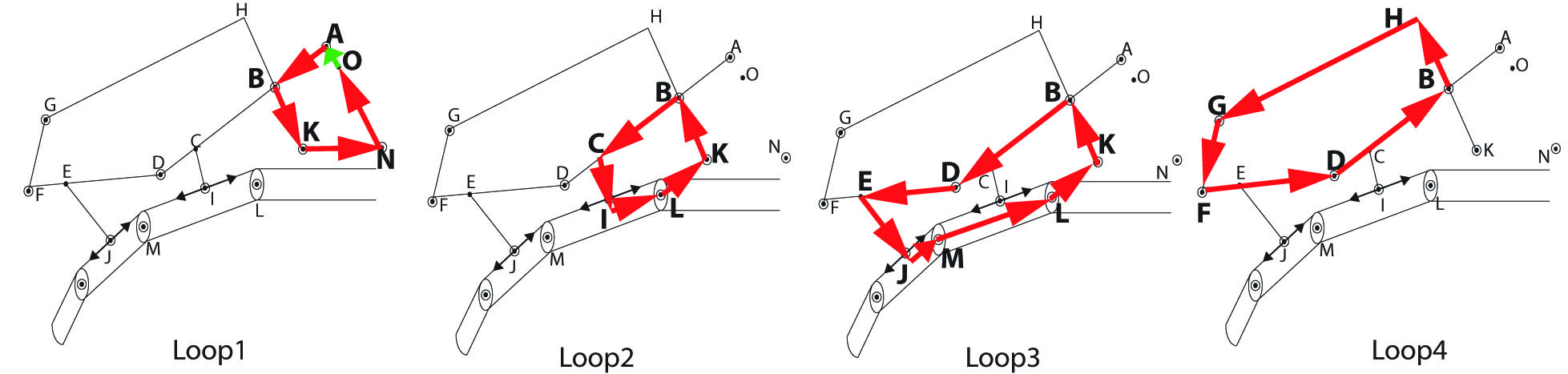}}
  %\vspace*{-0.5\baselineskip}
    \caption{Vector loops to analyze the kinematics of the underactuated hand exoskeleton.}
    \label{fig:loops}
    %\vspace*{-1\baselineskip}
\end{figure*}

Table~\ref{tab:variables} shows that the whole mechanism can be defined with the vectors using $8$ unknown parameters as \{$l_x$, $c_1$, $c_2$, $q_B$, $q_D$, $q_G$, $q_K$, $q_N$\} and the given finger pose variables as \{$q_{o1}$, $q_{o2}$\}. Even though there are other passive revolute joints that perform rotation during operation as \{$q_A$, $q_F$, $q_I$, $q_J$\}, their rotations are not required to define the device configuration since they are constrained by other variables stated in Table~\ref{tab:variables}. To find a unique solution for these $8$ unknowns, $8$ independent equations needed: four independent  loops were identified and are shown in Figure~\ref{fig:loops}, where  the vector chains along the mechanical links.

In particular, $Loop 1$ provides a relation between the actuator displacement and the mechanism itself using the vector loop equation and its corresponding exponential expression as

\begin{align}
\label{eq:loop1}
 \nonumber \vec{\vect{r}^{O}_{A} + \vect{r}^{A}_{B} + \vect{r}^{B}_{K} + \vect{r}^{K}_{N} + \vect{r}^{N}_{O} = \vect{0}} \\
  l_xe^{iq_N} + l_{AB}e^{iq_B} + l_{BK}e^{iq_K} + l_{KN}e^{iq_{KN}} + l_{act}e^{iq_N} = 0
\end{align}

\noindent where the unknown parameters observed from this loop can be listed as \{$l_x$, $q_B$, $q_K$, $q_N$\}. $Loop 2$ defines the corresponding motion around the MCP joint of the user's finger using the passive joints affected by the actuator.

\begin{align}
 \label{eq:loop2}
  \nonumber \vec{\vect{r}^{K}_{B} + \vect{r}^{B}_{C} +   \vect{r}^{C}_{I} + \vect{r}^{I}_{L} + \vect{r}^{L}_{K} =  \vect{0}}\\
  l_{BK}e^{iq_K} + l_{BC}e^{iq_B} + l_{CI}e^{iq_B} + c_1e^{iq_{o1}} + l_{LK}e^{iq_{LK}} = 0
\end{align}

\noindent The unknown parameters of $Loop 2$ can be listed as \{$q_K$, $q_B$, $c_1$\}, while also the finger pose parameter \{$q_{o1}$\} is included. Similarly, the PIP joint rotation of the user's finger is defined by $Loop 3$ using the passive joints of the mechanism.

\begin{align}
 \label{eq:loop3}
  \nonumber \vec{\vect{r}^{K}_{B} + \vect{r}^{B}_{D} +  \vect{r}^{D}_{E} + \vect{r}^{E}_{J} + \vect{r}^{J}_{M} + \vect{r}^{M}_{L} + \vect{r}^{L}_{K} =  \vect{0}} \\ \nonumber
  l_{BK}e^{iq_K} + l_{BD}e^{iq_B} + l_{DE}e^{iq_D} + l_{EJ}e^{iq_D} + c_2e^{iq_{o2}} \\
  + l_{ML}e^{iq_{o1}} + l_{LK}e^{iq_{LK}} = 0
\end{align}

\noindent The unknown and finger pose parameters are \{$q_D$, $c_2$\} and \{$q_{o2}$, $q_{o1}$\} respectively. Finally, $Loop 4$ provides a relation between the mechanical passive joints alone using the variables \{$q_K$, $q_B$, $q_G$, $q_D$\} in order to complete the number of equations to achieve a unique solution for the pose analysis.

\begin{align}
  \label{eq:loop4}
  \nonumber \vec{\vect{r}^{B}_{H} + \vect{r}^{H}_{G} +  \vect{r}^{G}_{F} + \vect{r}^{F}_{D} +  \vect{r}^{D}_{B} =  \vect{0}} \\
  l_{BH}e^{iq_K}+l_{HG}e^{iq_K}+l_{GF}e^{iq_G}+l_{FD}e^{iq_D}+l_{DB}e^{iq_K}=0
\end{align}

The loops above have been chosen such that one cannot be obtained using the others among $Loop 1$ - $Loop 4$. Furthermore, each loop provides at least one unique unknown parameter that is not covered by other loops as \{$l_x$, $q_N$ \} in $Loop 1$, \{$c_1$\} in $Loop 2$, \{$c_2$\} in $Loop 3$ and \{$q_G$\} in $Loop 4$. The existence of these unique parameters can be stated as a proof of independency of the loop equations. Even though the loops above can be chosen in a different manner as well, the simplest paths have been chosen to satisfy the purpose of the loop.

The $X$ and $Y$ components of the vector equations in Eqns.~(\ref{eq:loop1} - \ref{eq:loop4}) achieve $8$ nonlinear equations. The analytical solution cannot be achieved due to the nonlinearity of the system. Yet, the numerical methods gives a unique configuration solution for the actuator displacement $l_x$ as well as the passive joints [$c_1$, $c_2$, $q_B$, $q_D$, $q_G$, $q_K$, $q_N$] for given finger pose $q_{o1}$ and $q_{o2}$. These loop equations can also be used to calculate the Jacobian of the system.

\subsection{Differential Kinematics}

In order to calculate the differential kinematics of the system, $8$ nonlinear equations obtained by Eqns.~(\ref{eq:loop1} - \ref{eq:loop4}) should be differentiated. The purpose of the Jacobian is to obtain the angular velocities around the finger joints ($\dot{q}_{fin}$ = $\left[\dot{q}_{o1}, \dot{q}_{o2} \right]$). Even though the underactuated system provides a single actuator to control two joints ($l_x$), an additional joint is assumed to be measured to obtain the invertibility of the Jacobian ($q_B$). Therefore, the Jacobian should be calculated to obtain $\dot{q}_{fin}$ using the measured velocities ($\dot{q}_m$ = $\left[\dot{l}_x, \dot{q}_{B} \right]$) with the presence of passive velocities ($\dot{q}_p$ = $\left[\dot{q}_K, \dot{q}_{D}, \dot{q}_{G}, \dot{q}_{N}, \dot{c}_{1}, \dot{c}_{2} \right]$). Please note that the measured and passive velocities are the combination of linear and angular velocities. Taking the derivatives of the Eqns.~(\ref{eq:loop1} - \ref{eq:loop4}) can be categorized in the matrix form as in Eqn.~\ref{eq:matrix_def}.

\begin{equation} \label{eq:matrix_def}
  \left[
  \begin{array}{r}
  J_{O_{m}} \\
  J_{O_{p}}
  \end{array} \right]
  \vect{\dot{q}_{fin}}  = \left[
  \begin{array}{rr}
  J_{R_{m}} & J_{R_{p}} \\
  J_{C_{m}} & J_{C_{p}}
  \end{array} \right]
  \begin{bmatrix}
  \vect{\dot{q}_{m}} \\
  \vect{\dot{q}_{p}}
  \end{bmatrix}
\end{equation}

\noindent where $J_{O_m}, J_{O_p}, J_{T_{m}}, J_{T_{p}}, J_{C_{m}}, J_{C_{p}}$ are the matrices with the size of $2 \times2, 6 \times 2, 2 \times 2, 2 \times 6, 6 \times 2, 6\times6$ respectively. These matrices include the coefficients of the derivative terms coming from the 8 equations obtained by Eqns.~(\ref{eq:loop1} - \ref{eq:loop4}). In particular, the order of the equations have been chosen  such that none of the submatrix components are zero matrices, where all the elements are $0$, even though they can have $0$ value in their elements. In particular, the matrix components $J_{._m}$ indicate the coefficients regarding the measured variables while the components  $J_{._p}$  the passive variables. Similarly, $J_{O_.}$ indicates the output coefficients, $J_{C_.}$  the constraint coefficients and the terms $J_{T_.}$ are named after the total Jacobian terms. In order to simplify the Jacobian calculation of the overall system, $\dot{q}_p$ can be expressed in terms of other components using the second row of Eqn.~\ref{eq:matrix_def}

\begin{equation}
\vect{\dot{q}_p} = J^{-1}_{C_{p}} [J_{O_{p}} \vect{\dot{q}_{fin}} - J_{C_{m}} \vect{\dot{q}_{m}}]
\end{equation}

\noindent which can be used to replace the term $\dot{q}_p$ in the first row of Eqn.~\ref{eq:matrix_def} as

\begin{align} \label{eq:jacobian}
\nonumber \vect{\dot{q}_{fin}} &= [J_{O_m} - J_{T_{p}} J^{-1}_{C_{p}} J_{O_p}]^{-1} [J_{T_{m}} - J_{T_{p}} J^{-1}_{C_{p}} J_{C_{m}}] \vect{\dot{q}_{m}}
\\
 &= J_A \vect{\dot{q}_{m}}
\end{align}

The Jacobian obtained by Eqn.~\ref{eq:jacobian} is the relation between the measured and output velocities, while the inverse Jacobian transpose can be used to provide the torques applied to the finger joints for given actuator force.

\subsection{Statics Analysis and Stability of Grasp}\label{sec:statics}

The stability and the safety of the grasping tasks can be guaranteed only if the transmitted forces are applied in the correct direction to provide an interaction between the grasping object and the user's finger. Since the proposed underactuated mechanism does not control independently the value of  forces at the two phalanges, a static analysis is crucial to ensure the stability of the grasping tasks at any pose of the mechanism.

The static analysis to ensure the stability of the grasping forces has been already introduced previously  for a fully mechanical underactuated gripper~\cite{Gosselin2003} through the formulation $\vect{f_{fin}} = {J_{T}^{-T}} {J_{A}^{-T}} \vects{\tau_{m}}$, where $\vect{f_{fin}}$ is the vector of forces acting from the finger phalanges to the grasping object, $\vects{\tau_{m}}$ is the dual force vector of $\vect{\dot{q_m}}$, $ {J_{T}^{-T}}$ is the inverse transpose of the Jacobian between the angular velocities around the finger phalanges and the linear velocities at the contact points and ${J_{A}^{-T}}$ is the inverse transpose of the Jacobian calculated in Eqn.~\ref{eq:jacobian}. However, the static analysis in this work aims to control the force transmission on the phalanges in terms of finger joint torques $\vects{\tau_{fin}}=[\tau_1;  \tau_2]$ in order to optimize the link lengths of the mechanism. The force transmission is simplified and can be obtained as Eqn.~\ref{eq:static} using the inverse Jacobian transpose (${J_{A}^{-T}}$) as calculated in Eqn.~\ref{eq:jacobian}.

\begin{align}
\label{eq:static}
\vects{\tau_{fin}} = {J_{A}^{-T}} \vects{\tau_{m}}
\end{align}

\noindent where $\vects{\tau_{m}}=[f_{ac}; \tau_B]$, and since  it can be assumed that $\tau_B=0$, we obtain a direct relationship between the actuator force and the contact forces. The analysis of sign of $\tau_1$ and $\tau_2$ allows to study the stability of grasp under the actuator action while in contact with an object.

\section{Link Length Optimization} \label{sec:optimization}

A link length optimization was performed for each finger mechanism to improve the overall operation performance satisfying  the  following physical constraints:

\begin{itemize}
	\item The device is connected to the finger phalanges with passive linear sliders. Since these sliders have to be fitted on the finger phalanges, their movements ($c_1$ and $c_2$) have to be limited by the human finger phalanges measurements.
	\item The closing/opening of the hand is performed by the transmission of the forces to the finger joints (MCP and PIP). To provide a stable grasping, the two forces have to be balanced and should be always directed with the same sign (towards opening or closure of the hand).
	\item Since the analysis of the mechanism movement can be calculated only by given finger joints, the required actuator displacement should be limited by the choice of the linear actuator.
\end{itemize}

 The optimization has been conducted by an extensive search procedure  to find the  link length that maximize the following cost function $p$:

   \begin{numcases}  { \max~p = \sqrt{\tau_1^2 + \tau_2^2}, \text{ such that: }  \label{eq:opt}  }
   0\leq l_x\leq l_{max} \nonumber \\
   0\leq c_1\leq c_{1max} \nonumber \\
   0\leq c_2\leq c_{2max} \nonumber \\
   1\leq \tau_1/\tau_2 \leq 7.5 \nonumber
   \end{numcases}

  \noindent where $\tau_1$ and $\tau_2$ are defined as the torques acting on the MCP and PIP joints, $c_{1max}$ and $c_{2max}$ are determined as $50~mm$ and $40~mm$ and $l_{max}$ is $50~mm$ due to the actuator choice. The limitation on $c_{2max}$ violates the length of the second phalange, since the largest displacement occurs at this slider and there is no mechanical interaction by exceeding this finger phalange dimensions.

The search space is defined by the constant  link length parameters defined in the third column of table \ref{table_parameters}.
Before the optimization, a set of initial lengths were selected  to define a reasonable range of link lengths belonging to the search space, than the following computational steps were followed:

%If the current iteration does not satisfy any of these constraints, it is interrupted and the next iteration is initialized. Since the underactuation allows the device to cover different combinations of the first and second finger joint rotations, these constraints have to be satisfied for all possible configurations of the device. Therefore, an iteration is performed from $0^o$ to $80^o$ for MCP, $0^o$ to $90^o$ for MCP joints to cover the overall workspace without violating these constraints. The operational performance of the system is defined based on the force distribution of the mechanism from the actuator to the finger phalanges and finger joints. If the output link set, which is on the boundary of the defined range, the range should be extended if not bounded by a physical constraint. Doing so, the iteration based optimization can claim to reach a minimal solution for the defined performance objective..

\begin{enumerate}
  \item * Define a feasible and wide link length range
  \item Move the finger joints from $0^o$ to $80^o$ for MCP, $0^o$ to $90^o$ for MCP joints
  \item Compute the movement on the passive prismatic joints $c_1$, $c_2$ and actuator $l_x$
    \begin{itemize}
      \item Control if $l_x$, $c_1$ and $c_2$ satisfy the physical limits
      \item Go to the next set and start from 2 if not satisfied.
    \end{itemize}
  \item Compute the torques on the finger joints if $1~N.$ force is applied from the actuator using Jacobian transpose
    \begin{itemize}
     \item Control if the ratio between the torques of two joints satisfy the predefined limits
      \item Go to the next set and start from 2 if not satisfied.
    \end{itemize}
  \item Iterate the finger joints until MCP and PIP joints reach up to $80^o$ and $90^o$ respectively
    \begin{itemize}
      \item Repeat 3.
      \item Repeat 4.
    \end{itemize}
  \item Calculate the optimization objective
  \vspace*{.5\baselineskip}
    \begin{itemize}
      \item  $p = \sqrt{\tau_1^2 + \tau_2^2}$
    \end{itemize}
\end{enumerate}

 For the sake of simplifying the search space of the optimization, a sensitivity analysis was conducted to identify the link length  parameters that do not affect significantly the  optimization procedure and so reduce the dimension of the search space.

\subsection{Sensitivity Analysis}

 Although the numerical or analytical derivatives of the overall cost  function $p$ with respect to each search parameter provides an efficient approach,  the derivatives are not easy to obtain for  complex non linear models. The one-at-a-time (OAT) sensitivity analysis is an alternative method  analyzing the effect of a single parameter on a cost function, keeping the other parameters fixed~\cite{Mouida2011}. For this purpose, a sensitivity index ($SI$) is used as expressed in Eqn.~(\ref{eq:si}).

\begin{equation}\label{eq:si}
SI = \displaystyle \frac{\frac{S_2-S_1}{S_{av}}}{\frac{E_2-E_1}{E_{av}}}
\end{equation}

\noindent where $SI$ is the sensitivity index of the model output, $E_1$ and $E_2$ are the minimum and maximum values of the input parameters; $S_1$ and $S_2$ are the corresponding output values for $E_1$ and $E_2$; $S_{av}$ and $E_{av}$ are the average values of input and output parameters respectively. This index provides a quantitative relation between the model outputs and the input variables in terms of sensitivity. Negative $SI$ indicates that the inputs and outputs vary in opposite directions, while positive values signify a change in the same trend. In particular, the aim is to choose the positive, higher than unity $SI$ values, which show that a change in the parameter creates a higher effect in the output.

\begin{figure}[htb]
  \centering
     % \vspace*{-2\baselineskip}
  \resizebox{3.2in}{!}{\includegraphics{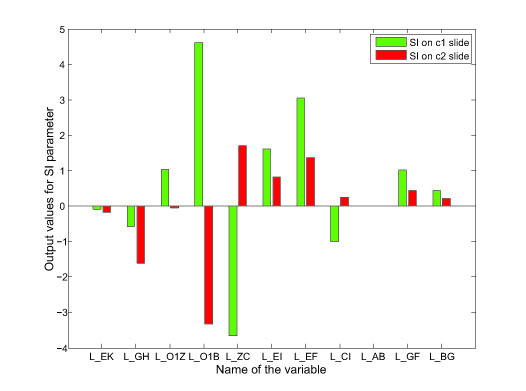}}
  \caption{Results of the sensitivity analysis for the $c_1$ and $c_2$ passive linear joints.}
  \label{fig:sensit}
  %  \vspace*{-1\baselineskip}
\end{figure}

While performing a closing/opening of the human finger, the most crucial output values have been stated as the displacement of the passive linear sliders ($c_1$ and $c_2$) due to the limitation imposed by the finger phalanges. Therefore, the effect of each variable has been investigated on these values individually. A representative   pose of the finger was used for this analysis. During the sensitivity analysis, each input variable was changed by $\pm 10\%$ from the initial value, keeping the other variables constant. For each set of lengths, the linear displacements, the variation of lengths $c_1$ and $c_2$, were calculated using the pose analysis as discussed in Section~\ref{sec:kinematics}. The sensitivity index has been calculated individually for $c_1$ and $c_2$ where the analysis limit is set to $+10\%$ for both cases. In other words, the length parameters, which result in negative or $\leq 0.1$ SI values, are not considered as effective variables over the performance of the linear movement. Fig.~\ref{fig:sensit} represents the sensitivity analysis results for the displacement of $c_1$ and $c_2$ individually, calculated as $SI_{c_1}$ and $SI_{c_2}$ individually. Note that the bars with the values under 0.001, cannot be seen in the plot, as for the $L_{AB}$.

It can be easily observed that for some variables, the sign of the SI values are different for $c_1$ and $c_2$ displacements signifying that the increase of the length affects the sliders in different ways. Eqn.~\ref{eq:si2} combines the SI values on both sliders where the positive values indicate the similar output effect on the sliders and the negative values indicate different behaviours. Table~\ref{tab:sens} presents the output values of  $SI_{g}$ from Eqn.~\ref{eq:si2}.

\begin{equation}\label{eq:si2}
SI_{g} = \text{sign}(SI_{c1})\cdot\text{sign}(SI_{c2}) \sqrt{SI_{c1}^2 + SI_{c2}^2}
\end{equation}

\begin{table}[htb]
    %\vspace*{-1\baselineskip}
\caption{Results of the generic sensitivity index.}
\label{tab:sens}
\begin{center}
\resizebox{2.5in}{!}{
\begin{tabular}{c c||c c}
  \hline \hline \\
  variable & value [mm] & variable & value [mm] \\
  \hline
  $SI_{EJ}$ & 0.1939 & $SI_{ED}$ & 3.3437 \\
  $SI_{CI}$ & 1.7077 & $SI_{GF}$ & -1.0275 \\
  $SI_{KH}$ & -1.0298 & $SI_{AB}$ & -2.1864e-09 \\
  $SI_{KB}$ & -5.6804 & $SI_{CD}$ & 1.1021 \\
  $SI_{GH}$ & -4.0324 & $SI_{BC}$ & 0.4877 \\
  $SI_{EF}$ & 1.8146 & & \\
  \hline \hline
  %\hline
\end{tabular}}
\end{center}
%\vspace*{-2\baselineskip}
\end{table}

From the sensitivity analysis, the most efficient variables are obtained as $L_{EJ}$, $L_{CI}$, $L_{EF}$, $L_{ED}$, $L_{CD}$ and $L_{BC}$ as in Table~\ref{tab:sens}. It is important to note that, the negative values represent different effects on $c_1$ and $c_2$ sliders as a result for increased link length. These lengths are  excluded from the parameter search space since minimizing two of the slides simultaneously is not feasible, keeping them constant during the optimization.  These constant values for the index finger have been set as $L_{KH} = 72$, $L_{KB} = 35$, $L_{GH} = 86$, $L_{AB} = 20$, and $L_{GF} = 36$ in $mm$. The search space for the optimization procedure has been reduced to $6$ variables, with ranges reported for the index finger in Table~\ref{tab:ranges}.

\subsection{Pre-optimization procedure based on linear and static constraints}

 A preliminary optimization procedure has been conducted to select the combinations of parameters satisfying the displacement and static constraints among the link lengths as ranged in Table~\ref{tab:ranges}. The variables have been iterated with a difference of $1~mm$ within the given range throughout the optimization. For the pre-optimization procedure, the MCP and PIP joints are moved in different paths for the corresponding iteration set to check whether the linear and static constraints are satisfied. The length combinations are eliminated from the optimization performance if the constraints above are not satisfied.

\begin{table}[H]
%\vspace*{-1\baselineskip}
\caption{Range of variables to be used for optimization.}
\label{tab:ranges}
\begin{center}
\resizebox{2.7in}{!}{
\begin{tabular}{c c||c c}
  \hline \hline \\
  variable & range [mm] & variable & range [mm] \\
  \hline
  $L_{EJ}$ & 30 - 40  & $L_{ED}$ & 30 - 40  \\
  $L_{CI}$ & 16 - 20  & $L_{EF}$ & 20 - 35  \\
  $L_{CD}$ & 10 - 20  & $L_{BC}$ & 36 - 46 \\
  \hline \hline
\end{tabular}}
\end{center}
%\vspace*{-2\baselineskip}
\end{table}

\subsubsection{Linear Constraints}

The passive linear constraints during the optimization aims to limit the passive linear movements, $c_1$ and $c_2$ calculated as in Subsection~\ref{sec:kinematics}, such that the passive cylindrical joints can be fitted on the finger phalanges. In fact, the linear movements should not exceed the average length of proximal and middle phalanges. The set of link lengths of the corresponding iteration is kept for the next phase of the optimization if the pose analysis using the constants of the current iteration results with the $c_1$ and $c_2$ outputs within the ranges of the phalange limits. Moreover, the required actuator displacement cannot exceed the properties of the chosen actuator, $50~mm.$ for this case. Within the given range, $65~\%$ of the iteration sets have been eliminated by the linear constraints of the index finger.

\subsubsection{Statical Constraints}

An iteration is used to calculate the corresponding torques applying on the finger MCP and PIP joints, $\tau_1$ and $\tau_2$ respectively in different configurations for $1~N$ applied force from the actuator, if the previous constraints have been satisfied. In fact, the calculation of $\tau_1$ and $\tau_2$ gives the same result using the statics as described previously and the Jacobian for a given orientation. The statical constraint has been set on the ratio between $\tau_1$ and $\tau_2$ such that the safety of the mechanical transmission can be investigated. In fact, this ratio, which has been calculated for different orientations of the device as $\tau_1/\tau_2$, is constrained by a minimum value of $1$ and the maximum value of $7.5$ without considering the contact forces. The statics constraint eliminates $90~\%$ of the remaining sets of variables for the index finger.

\subsection{Optimization}

To enlarge the efficient workspace of the finger joints, the previous constraints have been controlled again to explore a  workspace up to $80^o$ for the MCP joint and $90^o$ for the PIP joint. After the pre-optimization selection, where the physical constraints have been ensured to be satisfied, an optimization by exhaustive search has been conducted.

%The optimization objective has been defined to maximize the applied torques in the initial configuration of the exoskeleton where the achieved workspace is achieved as the maximum values of the finger joints as stated above. In fact the cost function $p$, which should be maximized, is stated as in Eqn.~(\ref{eqn:optimization}).

%\begin{equation}\label{eqn:optimization}
 % p = \sqrt{\tau_1^2 + \tau_2^2}
%\end{equation}

%\noindent where $\tau_1$ and $\tau_2$ are defined as the torques applied to the MCP and PIP joints.

\begin{figure}[htb]
  \centering
    % \vspace*{-2\baselineskip}
  \resizebox{3in}{!}{\includegraphics{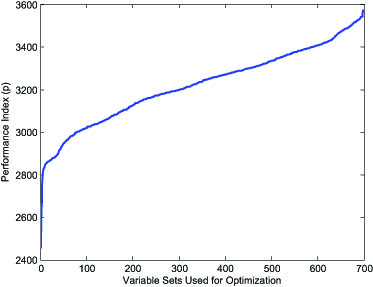}}
  \caption{Performance index $p$ as the sets of variable change. The order of these variable sets have been organized, such that the $p$ value is increasing to improve the visual impact of this image.}
  \label{fig:performance_plot}
    %\vspace*{-2\baselineskip}
\end{figure}

Each successful iteration that is not interrupted with the unsatisfied constraints is finalized by calculating the performance cost function. In order to provide a better understanding of the difference between the calculated cost function for each parameter set, Fig.~\ref{fig:performance_plot} shows the index $p$ calculation over the variable sets, which is ordered in a way to observe a constant increase. It can be observed that the choice of parameters can increase the performance by $\%50$. The variables are ordered by the performance index (p) to make it easier to visualize the change in the performance over the variable sets. Note that the range of variables for the links $L_{GH}$ and $L_{GF}$ could not be enlarged more to avoid any possible mechanical interference. The obtained RoM for the finger joints is effective to perform grasping tasks with real objects.

\begin{table}[htb]
    %\vspace*{-2\baselineskip}
\caption{Results of optimization.}
\label{tab:results}
\begin{center}
\resizebox{2.7in}{!}{
\begin{tabular}{c c||c c}
  \hline \hline \\
  variable & value [mm] & variable & value [mm] \\
  \hline
  $L_{EJ}$ & 37 & $L_{ED}$ & 32 \\
  $L_{CI}$ & 16 & $L_{EF}$ & 30 \\
  $L_{CD}$ & 10 & $L_{BC}$ & 42 \\
  \hline \hline
\end{tabular}}
\end{center}
%\vspace*{-2\baselineskip}
\end{table}

\begin{figure}[htb]
  \centering
     % \vspace*{-1\baselineskip}
  \resizebox{3.4in}{!}{\includegraphics{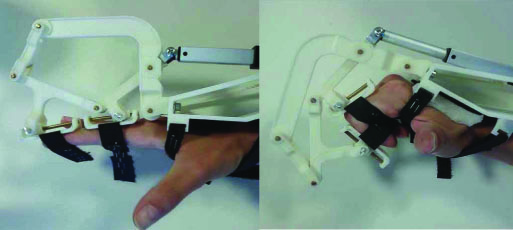}}
  \caption{Maximum flexion and extension of the index finger, while the user is wearing a finger component of the underactuated hand exoskeleton.}
  \label{fig:optimal_solution}
  %  \vspace*{-1\baselineskip}
\end{figure}

For the middle-sized index finger, the optimized link lengths allow the finger joints to reach the workspace stated in Table~\ref{tab:index_rom_exo}. Furthermore, Figure~\ref{fig:optimal_solution} shows the finger exoskeleton, which have been manufactured by a 3-D printer, in maximum flexion and extension configurations of the index finger of a user.

\begin{table}[htb]
     %\vspace*{-2\baselineskip}
\caption{Ranges of Motion for Finger Joints with Hand Exoskeleton}
\label{tab:index_rom_exo}
\begin{center}
\begin{tabular}{c||c|c|c}
\hline \hline \\
  Joint & MCP & PIP & DIP \\
  RoM & 0 - 80 & 0 - 90 & -- \\
  \hline \hline
\end{tabular}
\end{center}
    %\vspace*{-2\baselineskip}
\end{table}

The device is optimized considering the measurements of the middle-sized hand, where the index finger has the measurements of $50~mm$ and $30~mm$ for the first and second finger phalanges. However, it is observed that the device with selected link lengths satisfy the physical constraints and provide an efficient operation also for the small-sized hand (index finger of $45~mm$, $27~mm$ respectively) and big-sized hand (index finger of $55~mm$, $34~mm$ respectively), which cover a wide range of users. Still, it is important to note that the ranges of motion for finger joints stated in Table~\ref{tab:index_rom_exo} tend to  change for different hand sizes, since the movements change and the movement is mostly limited by the physical workspace of the passive prismatic joints $c1$ and $c2$.

The initial choice of the link $L_{KH}$ increases the overall mechanism bulk, but should be determined  to prevent a mechanical interference during operation.

% The large space might be seen as an encumbrance of the device to limit its range of applicability and its usability during some tasks. However, the 3D printed mechanical parts allow the device to be lightweight not to cause fatigue during the experiments. Furthermore, connecting the device to the user’s hand in a stable manner prevents the device to unbalance during the operation. With this motivation, we claim that this device can provide an effective use for haptic or rehabilitative applications.

\section{Multi-Finger Exoskeleton}\label{sec:hand}

The optimization procedure has been repeated for each finger to optimize the link lengths and to achieve a multi-finger exoskeleton. The first prototype of the hand exoskeleton has been made with rapid prototyping parts, allowing the device to be low-cost and light-weight. For the actuation, Firgelli $L16$ linear motors have been used with $50~mm$ linear stroke for each finger thanks to their low cost and high availability. Moreover, the small case and the low-weight of the actuators allow the placement on top of the hand without causing fatigue during operation. Table~\ref{tab:properties} presents the mechanical specifications of the device and the actuators.

\begin{table}[htb]
    %\vspace*{-1\baselineskip}
\caption{Specifications of the proposed exoskeleton and its actuators.}
\label{tab:properties}
\begin{center}
\resizebox{2.7in}{!}{
\begin{tabular}{l||c}
  \hline \hline \\
  {\bf property} & value [mm]\\
  \hline
  Device Mass & $\cong$ 300 g.\\
  RoM for MCP & $80^o$  \\
  RoM for PIP & $90^o$   \\
  Motor gear ratio & 35:1 \\
  Stroke of the motor & 50 mm \\
  Max. cont force of the motor & 40 N  \\
  Max. torque on MCP & 1485 Nmm\\
  Max. torque on PIP & 434 Nmm\\
  Backdrive force & 31 N  \\
  Max. velocity &  32 mm/s \\
  \hline \hline
\end{tabular}}
\end{center}
%\vspace*{-1\baselineskip}
\end{table}

\begin{figure}[htb]
  \centering
      %\vspace*{-1\baselineskip}
  \resizebox{2.5in}{!}{\includegraphics{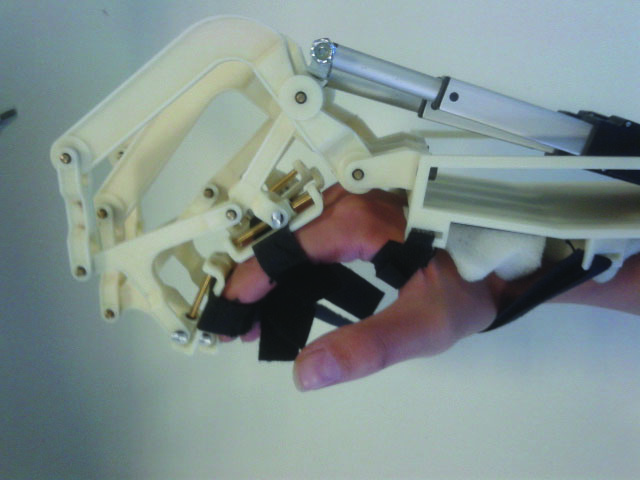}}
  \caption{Implementation of the multi-finger hand exoskeleton.}
  \label{fig:multiCAD}
    %\vspace*{-1\baselineskip}
\end{figure}

The straps around the user's hand and the finger phalanges allow the device to be worn in about a minute, without any initial pose requirement of the human finger. Figure~\ref{fig:multiCAD} shows the hand exoskeleton that is connected to the index, middle and ring fingers individually. The thumb and the little fingers are more complicated due to their size and the complexity of design, so they are left as a future work. During the grasping tasks, the thumb is left passive and the little finger is attached to the ring finger from the proximal phalanges by the straps.

\begin{figure}[htb]
  \centering
      %\vspace*{-1\baselineskip}
  %\resizebox{2.3in}{!}{\includegraphics{grasping_hand}}
  \resizebox{3in}{!}{\includegraphics{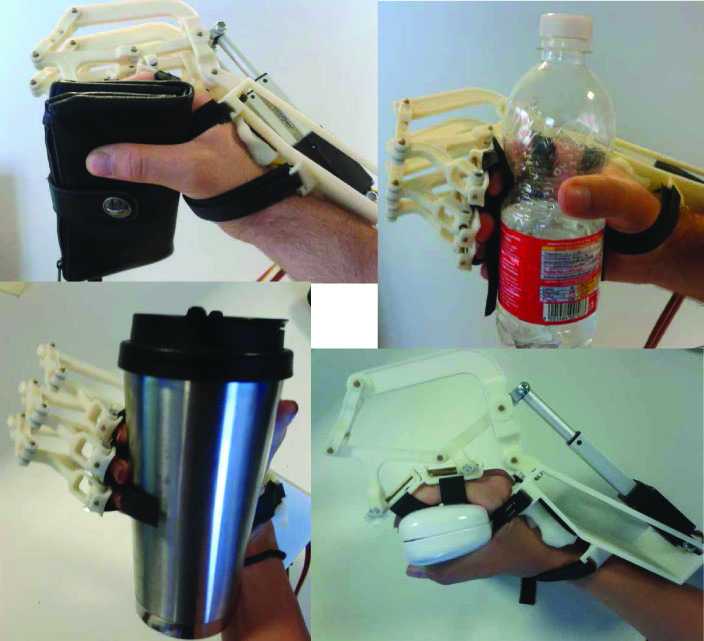}}
  \caption{Underactuated hand exoskeleton with multiple finger components can assist users to grasp different objects with random sizes and random shapes.}
  \label{fig:grasping_hand}
   % \vspace*{-1\baselineskip}
\end{figure}

The usability of the device has been tested through the grasping tasks of various objects and the force transmission on the finger phalanges to perform the grasping. The mechanism is controlled by a simple position control.
Figure~\ref{fig:grasping_hand} shows that the users can perform the grasping of objects with different sizes and shapes with no preliminary knowledge of the object. Note that the first two tasks have been completed by a male user and the last two tasks have been completed by a female user. The adaptation of the device for two different size hands automatically is observed.

\begin{figure}[htb]
  \centering
      %\vspace*{-2\baselineskip}
  %\resizebox{2.3in}{!}{\includegraphics{grasping_hand}}
  \resizebox{3.2in}{!}{\includegraphics{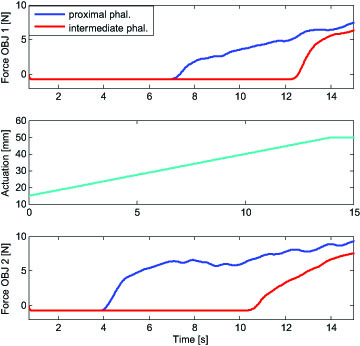}}
  \caption{Grasping forces have been collected on the finger phalanges of a user, while the user is grasping different objects to validate the power grasping of the hand exoskeleton.}
  \label{fig:grasping_force}
   %\vspace*{-1\baselineskip}
\end{figure}

For further analysis, the grasping forces with the assistance of the exoskeleton have been analyzed. The grasping forces applied by the proximal and intermediate phalanges of the index finger have been measured for two objects with different dimensions. The grasping forces are measured by two force sensitive resistors that have been attached on the contact points between the object and the user. The force sensor has been calibrated to obtain linear measurements up to $15N$ to ensure that the applied forces can be obtained efficiently. Figure~\ref{fig:grasping_force} shows the grasping forces over time for the given actuator displacements.

Grasping the first object, the actuator displacement is used to move the proximal phalange to the contact point with the object. The contact has been achieved when the force measurements of the proximal phalange starts to increase. Meanwhile, the movement of the actuator is transmitted to the intermediate phalange until it reaches the object as well. The timings and the values of force measurements for each object show difference to indicate different behaviour for grasping, which is adjusted automatically by the kinematics of the device. Yet, the stable grasping can be achieved for both cases. Even though the grasping forces are shown only for the index finger phalanges, other fingers have been observed to have the similar behavior as well.

\section{Conclusion and Future Work} \label{sec:conc}

In this paper, we have presented a novel, underactuated finger exoskeleton that exerts only normal forces to the finger phalanges while performing flexion/extension movements for grasping in a natural and safe manner. The kinematic synthesis of the linkage based design has been performed to achieve all the proposed requirements, and the pose of the device has been solved for given finger configuration. The statics of the mechanism for different configurations of the finger was derived to verify the stability of grasping. Finally, an optimization procedure has been defined to determine the link lengths that maximize the force transmission along the mechanical links, while satisfying the kinematic and statical constraints for the correct working of the device and reaching a sufficient range of motion to perform grasping tasks.

The first prototype of the hand exoskeleton has been manufactured for index, middle and ring fingers while the thumb and the little finger implementations have been left as a future work. The preliminary tests show that the device can be adjusted to the users with different hand sizes. Moreover, the underactuation of the device has been tested by grasping different objects with various size and shape. The measurements on the finger phalanges during grasping tasks show that the stable grasping can be achieved with no prior information of the grasping object, thanks to the underactuation.

In the future, the hand exoskeleton will be completed by implementing the absent fingers. The backdriveability is aimed to be achieved by control to be able to move the exoskeleton freely while attached to the user's hand. Regarding the adaptability of the device by different hand sizes, further experiments are required to understand the achieved workspace by each user. In case of need, multiple sizes of the device can be manufactured to target small, medium or big hand sizes and minimize the output performance between hand sizes.

\vspace{-2mm}
\section{Acknowledgements}
This research was funded within the project "WEARHAP – WEARable HAPtics for humans and robots" of the European Union Seventh Framework Programme FP7/2007-2013, grant agreement n. 601165 and project
"CENTAURO - Robust Mobility and Dexterous Manipulation in Disaster Response by Fullbody Telepresence in a Centaur-like Robot"  of the   the European Union's Horizon 2020 Programme, Grant Agreement  n. 644839.
% \vspace{-1mm}

\small
\bibliographystyle{IEEEtran}
%%\vspace{-1mm}
%%\bibliographystyle{unsrt}
 \bibliography{HandExos_Revisionss_v3}

\begin{thebibliography}{10}
\providecommand{\url}[1]{#1}
\csname url@samestyle\endcsname
\providecommand{\newblock}{\relax}
\providecommand{\bibinfo}[2]{#2}
\providecommand{\BIBentrySTDinterwordspacing}{\spaceskip=0pt\relax}
\providecommand{\BIBentryALTinterwordstretchfactor}{4}
\providecommand{\BIBentryALTinterwordspacing}{\spaceskip=\fontdimen2\font plus
\BIBentryALTinterwordstretchfactor\fontdimen3\font minus
  \fontdimen4\font\relax}
\providecommand{\BIBforeignlanguage}[2]{{%
\expandafter\ifx\csname l@#1\endcsname\relax
\typeout{** WARNING: IEEEtran.bst: No hyphenation pattern has been}%
\typeout{** loaded for the language `#1'. Using the pattern for}%
\typeout{** the default language instead.}%
\else
\language=\csname l@#1\endcsname
\fi
#2}}
\providecommand{\BIBdecl}{\relax}
\BIBdecl

\bibitem{Kobayashi2012}
F.~Kobayashi, G.~Ikai, W.~Fukui, H.~Nakamoto, and F.~Kojima, ``Multipoint
  haptic device for robot hand teleoperation,'' in \emph{International
  Symposium on Micro-Nano Mechatronics and Human Science ({MHS})}, 2012, pp.
  304--309.

\bibitem{Lee2013}
S.~W. Lee, K.~Landers, and H.-S. Park, ``Biomimetic hand exotendon device
  ({BiomHED}) for functional hand rehabilitation in stroke,'' in \emph{IEEE
  International Conference on Rehabilitation Robotics ({ICORR})}, 2013, pp.
  1--4.

\bibitem{Brokaw2011}
E.~Brokaw, I.~Black, R.~Holley, and P.~Lum, ``Hand spring operated movement
  enhancer ({HandSOME}): A portable, passive hand exoskeleton for stroke
  rehabilitation,'' \emph{IEEE Transactions on Neural Systems and
  Rehabilitation Engineering}, vol.~19, no.~4, pp. 391--399, 2011.

\bibitem{Aubin2013}
P.~Aubin, H.~Sallum, C.~Walsh, L.~Stirling, and A.~Correia, ``A pediatric
  robotic thumb exoskeleton for at-home rehabilitation: The isolated orthosis
  for thumb actuation ({IOTA}),'' in \emph{IEEE International Conference on
  Rehabilitation Robotics ({ICORR})}, 2013, pp. 1--6.

\bibitem{Li2011}
J.~Li, R.~Zheng, Y.~Zhang, and J.~Yao, ``{iHandRehab}: An interactive hand
  exoskeleton for active and passive rehabilitation,'' in \emph{IEEE
  International Conference on Rehabilitation Robotics (ICORR)}, 2011, pp. 1 --
  6.

\bibitem{Hasegawa2011}
Y.~Hasegawa, J.~Tokita, K.~Kamibayashi, and Y.~Sankai, ``Evaluation of
  fingertip force accuracy in different support conditions of exoskeleton,'' in
  \emph{IEEE International Conference on Robotics and Automation (ICRA)}, 2011,
  pp. 680--685.

\bibitem{Jones2012}
C.~Jones, F.~Wang, R.~Morrison, N.~Sarkar, and D.~Kamper, ``Design and
  development of the cable actuated finger exoskeleton for hand rehabilitation
  following stroke,'' \emph{IEEE/ASME Transactions on Mechatronics}, vol.~19,
  no.~1, pp. 131--140, 2014.

\bibitem{Iqbal2011}
J.~Iqbal, N.~Tsagarakis, and D.~Caldwell, ``A multi-{DoF} robotic exoskeleton
  interface for hand motion assistance,'' in \emph{IEEE International
  Conference of the Engineering in Medicine and Biology Society}, 2011, pp.
  1575--1578.

\bibitem{Tang2013}
Z.~Tang, S.~Sugano, and H.~Iwata, ``A finger exoskeleton for rehabilitation and
  brain image study,'' in \emph{IEEE International Conference on Rehabilitation
  Robotics (ICORR)}, 2013, pp. 1--6.

\bibitem{Taheri2014}
H.~Taheri, J.~B. Rowe, D.~Gardner, V.~Chan, K.~Gray, C.~Bower, D.~J.
  Reinkensmeyer, and E.~T. Wolbrecht, ``{Design and preliminary evaluation of
  the FINGER rehabilitation robot: Controlling challenge and quantifying finger
  individuation during musical computer game play},'' \emph{Journal of
  NeuroEngineering and Rehabilitation}, vol.~11, no.~10, pp. 1--17, 2014.

\bibitem{Chiri2012}
A.~Chiri, N.~Vitiello, F.~Giovacchini, S.~Roccella, F.~Vecchi, and M.~Carrozza,
  ``Mechatronic design and characterization of the index finger module of a
  hand exoskeleton for post-stroke rehabilitation,'' \emph{IEEE/ASME
  Transactions on Mechatronics}, vol.~17, no.~5, pp. 884--894, 2012.

\bibitem{Gosselin2003}
T.~Laliberte, L.~Birglen, and C.~Gosselin, ``{Underactuation in robotic
  grasping hands},'' \emph{Machine Intelligence \& Robotic Control}, vol.~4,
  no.~3, pp. 1--11, 2002.

\bibitem{Ertas2014}
I.~H. Ertas, E.~Hocaoglu, and V.~Patoglu, ``{AssistOn-Finger}: An
  under-actuated finger exoskeleton for robot-assisted tendon therapy,''
  \emph{Robotica}, 2014.

\bibitem{birglen2004kinetostatic}
L.~Birglen and C.~M. Gosselin, ``Kinetostatic analysis of underactuated
  fingers,'' \emph{IEEE Transactions on Robotics and Automation}, vol.~20,
  no.~2, pp. 211--221, 2004.

\bibitem{Wang2009}
J.~Wang, J.~Li, Y.~Zhang, and S.~Wang, ``Design of an exoskeleton for index
  finger rehabilitation,'' in \emph{Annual International Conference in Medicine
  and Biology Society ({EMBC})}, 2009, pp. 5957--5960.

\bibitem{leonardis2015emg}
D.~Leonardis, M.~Barsotti, C.~Loconsole, M.~Solazzi, M.~Troncossi, C.~Mazzotti,
  V.~Parenti~Castelli, C.~Procopio, G.~Lamola, C.~Chisari \emph{et~al.}, ``An
  emg-controlled robotic hand exoskeleton for bilateral rehabilitation,'' 2015.

\bibitem{Mouida2011}
A.~Mouida and N.~Alaa, ``Sensitivity analysis of {TSEB} model by
  {One-Factor-At-A-Time} in irrigated olive orchard,'' \emph{IJCSI
  International Journal of Computer Science Issues}, vol.~8, no.~1, pp.
  369--378, 2011.

\end{thebibliography}
\end{document}